\documentclass[11pt]{article}

\usepackage[preprint]{acl}

\usepackage{times}
\usepackage{latexsym}
\usepackage{longtable}
\usepackage{multirow}
\usepackage{graphicx}
\usepackage[T1]{fontenc}

\usepackage[utf8]{inputenc}

\usepackage{microtype}

\usepackage{inconsolata}

\usepackage{graphicx}
\usepackage{booktabs}
\usepackage{longtable}
\usepackage{enumitem}
\usepackage{listings}
\usepackage{amsmath}
\usepackage{amssymb}

\lstdefinelanguage{json}{
    basicstyle=\normalfont\ttfamily,
    showstringspaces=false,
    breaklines=true,
    literate=
     *{0}{{{\color{blue}0}}}{1}
      {1}{{{\color{blue}1}}}{1}
      {2}{{{\color{blue}2}}}{1}
      {3}{{{\color{blue}3}}}{1}
      {4}{{{\color{blue}4}}}{1}
      {5}{{{\color{blue}5}}}{1}
      {6}{{{\color{blue}6}}}{1}
      {7}{{{\color{blue}7}}}{1}
      {8}{{{\color{blue}8}}}{1}
      {9}{{{\color{blue}9}}}{1}
      {:}{{{\color{red}{:}}}}{1}
      {,}{{{\color{red}{,}}}}{1}
      {\{}{{{\color{black}{\{}}}}{1}
      {\}}{{{\color{black}{\}}}}}{1}
      {[}{{{\color{black}{[}}}}{1}
      {]}{{{\color{black}{]}}}}{1},
}

\title{Multiperspectivity as a Resource for Narrative Similarity Prediction}

\author{
 \textbf{Max Upravitelev\textsuperscript{1,2}},
 \textbf{Veronika Solopova\textsuperscript{1,2}},
 \textbf{Jing Yang\textsuperscript{1,2,3}},
\\
 \textbf{Charlott Jakob\textsuperscript{1,2}},
 \textbf{Premtim Sahitaj\textsuperscript{1,2}},
 \textbf{Ariana Sahitaj\textsuperscript{1,2}},
 \textbf{and Vera Schmitt\textsuperscript{1,2,3,4}}
\\
  \textsuperscript{1}Technische Universit\``at Berlin \\
  \textsuperscript{2}German Research Center for Artificial Intelligence (DFKI) \\
  \textsuperscript{3}BIFOLD – Berlin Institute for the Foundations of Learning and Data \\
  \textsuperscript{4}Centre for European Research in Trusted AI (CERTAIN) \\
\small{
   \textbf{Correspondence:} \href{max.upravitelev@tu-berlin.de}{max.upravitelev@tu-berlin.de}
 }
}

\begin{document}
\maketitle
% ...




\begin{abstract}
Predicting narrative similarity can be understood as an inherently interpretive task: different, equally valid readings of the same text can produce divergent interpretations and thus different similarity judgments, posing a fundamental challenge for semantic evaluation benchmarks that encode a single ground truth. Rather than treating this multiperspectivity as a challenge to overcome, we propose to incorporate it in the decision making process of predictive systems. To explore this strategy, we created an ensemble of 31 LLM personas. These range from practitioners following interpretive frameworks to more intuitive, lay-style characters. 
Our experiments were conducted on the SemEval-2026 Task 4 dataset, where the system ranked 13th out of 47 teams and achieved an accuracy score of 0.705. Accuracy improves with ensemble size, consistent with Condorcet Jury Theorem-like dynamics under weakened independence. Practitioner personas perform worse individually but produce less correlated errors, yielding larger ensemble gains under majority voting. Our error analysis reveals a consistent negative association between gender-focused interpretive vocabulary and accuracy across all persona categories, suggesting either attention to dimensions not relevant for the benchmark or valid interpretations absent from the ground truth. This finding underscores the need for evaluation frameworks that account for interpretive plurality.

\end{abstract}

\section{Introduction}

Narrative understanding has traditionally been studied in interpretive sciences, where meaning is often modeled as perspective-dependent and shaped by subjective interpretation. As \citet{Kommers2025} emphasize, interpretations can conflict and even contradict each other while remaining valid, thus posing a fundamental challenge for computational approaches. This is especially relevant in the context of semantic evaluation and multiperspectivity: any benchmarks targeting interpretive tasks are subject to encoding the particular interpretive perspectives of their creators and annotators. Hence, predicting narrative similarity, recently formalized as SemEval-2026 Task 4 by \citet{hatzel2026} and extending earlier works such as \citet{hatzel-biemann-2024-story, akter2024fansfacetbasednarrativesimilarity}, is a task situated at the intersection of computation and interpretation. 

In the following, we argue that multiperspectivity should be operationalized as a modeling component rather than treated as an evaluation artifact in narrative similarity prediction. This strategy is motivated by the Condorcet Jury Theorem (CJT), which states that collective decisions improve as independently competent voters are added \cite{shteingart2020}. While its formal independence assumption is violated in LLM-based ensembles, the theorem motivates the hypothesis that aggregating diverse interpretive perspectives can improve prediction quality. Though this principle has yielded performance gains in NLP tasks like sentiment analysis \cite{barcena2024application}, results across domains remain mixed \cite{lefort2024}. Building on this observation, we construct an ensemble of 31 LLM personas divided into Practitioners [P] (personas with explicit analytical frameworks such as critical hermeneutics, feminist literary criticism or postcolonial theory) and Lay People [L] (more intuitive ``everyday'' characters). 
We systematically investigate how ensemble size, diversity, and persona category relate to prediction accuracy, and find that the relation between individual persona performance and collective decision quality reveals patterns about the role of interpretive diversity in computational narrative understanding.

Our error analysis further reveals a systematic pattern that bears directly on the relationship between multiperspectivity and semantic evaluation benchmarks: when personas across all categories employ vocabulary distinctive of gender-focused and feminist interpretation (terms such as \textit{gender roles}, \textit{female protagonist}, or \textit{patriarchal}) their accuracy tends to drop. This finding allows for two complementary readings: Gender-related vocabulary may signal attention to narrative dimensions that, while interpretively valid, are simply not predictive of the similarity judgments encoded in the ground truth of the given benchmark. Alternatively, it may indicate that some perspectives produce valid but unrepresented interpretations, raising the question of how computational benchmarks should account for multiperspectivity in annotation and evaluation.

\section{Preliminaries and Related Work}

\paragraph{Narrative Similarity}

Computational approaches to narrative similarity span embedding-based methods for story reformulations \cite{hatzel-biemann-2024-story}, structural decomposition via narrative theory \cite{chun-2024-aistorysimilarity}, facet-based metrics grounded in 5W1H dimensions \cite{akter2024fansfacetbasednarrativesimilarity}, and claim-level distillation for tracking ideas across media discourse \cite{waight-etal-2025-quantifying}. Narrative similarity also has applications in disinformation analysis, from taxonomy-based grouping of propaganda narratives \cite{2025pn} by similarity to comparing similar features across individual instances \cite{solopova-etal-2024-check}.


\paragraph{Multi-agent Debating Systems}
While multi-agent debate (MAD) has become a prominent strategy for improving LLM reasoning and decision-making through iterative inter-agent interaction \citep{liang-etal-2024-encouraging,srivastava-etal-2025-debate,
han-etal-2025-debate}, recent work has shown that majority voting alone often captures most of the associated performance gains across a range of knowledge and reasoning benchmarks \citep{choi-etal-2025-debate,kaesberg-etal-2025-voting, zhu-etal-2026-demystifying}. Building on this insight, we investigate multi-persona ensemble voting for narrative similarity assessment, while leaving the comparison with iterative debate protocols to future work. To the best of our knowledge, this is the first work to employ diverse LLM persona ensembles with majority voting grounded in distinct interpretive perspectives for narrative similarity prediction.

\section{Methodology}

\paragraph{Narrative Similarity Task}

Given a triplet of fictional texts in English consisting of an \textit{Anchor} story as well as two candidate stories \textit{A} and \textit{B}, it should be decided which candidate is more similar to the anchor. The provided dataset is split into a development (dev) split with 200 samples and a test split with 400 samples, to be evaluated by measuring the accuracy of the predicted binary choice against gold data.

\paragraph{Crafting LLM Personas}


\begin{table}[t]
\small
\centering
\resizebox{\columnwidth}{!}{%
\begin{tabular}{p{2.0cm} p{1cm} p{4.0cm}}
\toprule
\textbf{LLM Persona} & \textbf{Role} & \textbf{System Prompt} \\
\midrule

Literary Critic & Practi-tioner & You are a Literary Critic analyzing narrative similarity. \\
Computational Narratologist & Practi-tioner & You are a Computational Narratologist. You excel at structural analysis of text and the structured extraction of narrative features, such as actors, relations, events, topics, themes, sentiments etc. \\
... & ... & ... \\
Postcolonial Critic & Practi-tioner & You are a Postcolonial Critic examining the texts in terms of global power dynamics. \\
Feminist Literary Critic & Practi-tioner & You are a Feminist Literary Critic examining gender-related dynamics within texts. \\
Gender Perspective Analyst & Practi-tioner & You are a Gender Perspective Analyst examining gender-related dynamics. \\
... & ... & ... \\
\midrule
High School Student & Lay & You are an 8th grader. You are not the best student, but you are doing alright. \\
Football Player & Lay & You are a professional football player and not sure why you got this task, but provide your perspective anyway. \\
... & ... & ... \\

\bottomrule
\end{tabular}%
}
\caption{Selected LLM Personas, full list can be found in Appendix \ref{tab:personas_full}}

\label{tab:personas_minimal}

\end{table}

Since we want to investigate how the ensemble size and ensemble diversity relate to performance on the accuracy metric, we create 31 LLM personas and run different experiments on the dev set to investigate majority voting behavior in order to identify configurations that maximize accuracy on both the dev and the test set. Our ensemble consists of two categories of LLM personas, which we partly created manually and partly generated and refined: Practitioners [P], which are personas with concrete interpretive frameworks for handling text and text analysis, and Lay People [L], a group of more intuitive ``everyday characters'' inspired by findings from \citet{kim-etal-2025-persona}, who showed that more complicated roles can lead to performance decreases when compared to simpler personas. 
Examples of our LLM personas can be found in Table \ref{tab:personas_minimal}. Every persona is essentially an LLM system prompt (like ``You are a Computational Narratologist. You excel at structural analysis of text and the structured extraction of narrative features, such as actors, relations, events, topics, themes, sentiments etc.''), concatenated with base instructions (documented in Appendix \ref{sec:appendix_prompts}).

Expanding on the idea of multiperspectivity, we also deploy multiple models for each prediction and per persona. We chose three open-weight models from three different model families: Gemma 27-b it \cite{gemmateam2025gemma3technicalreport}, Qwen3-14B \cite{yang2025qwen3technicalreport} and gpt-oss-20b \cite{openai2025gptoss120bgptoss20bmodel}.

\paragraph{Voting Configurations}
Each persona produces a single prediction per item on each model, yielding three levels of aggregation: (1)~\textit{Individual persona}: the prediction of one persona on one model, with no voting involved; (2)~\textit{Model-specific majority vote}: all 31~personas on a single model vote by simple majority; and (3)~\textit{Cross-model majority vote}: all 93 persona\,\texttimes\,model predictions (31~personas $\times$ 3~models) are combined into a single majority vote.

\section{Results}

\paragraph{Ensemble Size}


\setlength{\tabcolsep}{3pt} 
\renewcommand{\arraystretch}{0.96} 

\begin{table}[ht]
\centering
\footnotesize
\resizebox{\columnwidth}{!}{%
\begin{tabular}{@{} c c c c c @{}}
\toprule
Ensemble Size & Qwen3 14B & Gemma 3 27B-it & gpt-oss-20b & All \\
\midrule
\multicolumn{5}{@{}l}{\textbf{Majority Vote Accuracy}} \\
E=1   & 69.1±0.4 & 71.9±0.4 & 57.0±0.7 & 66.0±0.3 \\
E=3   & 71.7±0.5 & 73.4±0.4 & 58.4±0.9 & 69.8±0.4 \\
E=5   & 72.7±0.6 & 73.8±0.5 & 59.0±0.9 & 71.5±0.4 \\
E=10  & 73.8±0.7 & 73.9±0.4 & 59.8±1.1 & 72.9±0.6 \\
E=20  & 74.2±0.7 & 74.2±0.5 & 60.2±1.4 & 74.5±0.6 \\
E=30  & 74.4±0.8 & 74.4±0.7 & 60.5±1.6 & 75.0±0.7 \\
E=31  & 74.3±0.9 & 74.5±0.8 & 60.8±1.9 & 75.2±0.6 \\
\midrule
\multicolumn{5}{l}{\textbf{Oracle $K \geq 1$ Accuracy}} \\
E=1   & 68.6±0.4 & 71.9±0.4 & 56.4±0.7 & 65.6±0.3 \\
E=3   & 87.5±0.4 & 84.9±0.4 & 84.6±0.6 & 90.4±0.3 \\
E=5   & 92.3±0.4 & 88.7±0.4 & 92.5±0.5 & 95.8±0.2 \\
E=10  & 96.3±0.3 & 92.9±0.5 & 97.8±0.3 & 98.8±0.1 \\
E=20  & 98.4±0.3 & 95.9±0.8 & 99.5±0.2 & 99.8±0.1 \\
E=30  & 99.1±0.4 & 97.1±1.0 & 99.8±0.2 & 99.9±0.1 \\
E=31  & 99.1±0.4 & 97.2±1.0 & 99.8±0.2 & 99.9±0.1 \\
\bottomrule
\end{tabular}%
}
\caption{Ensemble accuracy (\% ± standard deviation, based on random persona combinations with maximum sample limit of 5000.) by ensemble size, averaged over $n=10$ runs.}
\label{tab:acc_enseble_size}
\end{table}

To investigate the effects of ensemble size, we run a series of experiments based on different size settings. The results are documented in Table~\ref{tab:acc_enseble_size}. The upper block confirms that the accuracy of the majority vote increases with ensemble size $E$. The lower block reports an oracle analysis: given access to the ground truth, we measure the proportion of items for which at least $K \geq 1$ members of the ensemble yield the correct prediction. Here, a similar behavior can be observed where the results plateau around $E=30$.

\begin{table}[ht]
\centering
\resizebox{\columnwidth}{!}{%
\begin{tabular}{lccc}
\toprule
Metric & Qwen3 14B & Gemma 3 27B-it & gpt-oss-20b \\
\midrule
\(K\ge1\)   & \(99.1\%\pm0.4\%\)  & \(97.2\%\pm1.0\%\)  & \(99.8\%\pm0.2\%\)  \\
\(K\ge2\)   & \(97.8\%\pm0.5\%\)  & \(94.5\%\pm0.8\%\)  & \(99.3\%\pm0.5\%\)  \\
\(K\ge3\)   & \(96.4\%\pm0.5\%\)  & \(92.5\%\pm0.7\%\)  & \(98.5\%\pm0.7\%\)  \\
\(K\ge4\)   & \(95.0\%\pm0.5\%\)  & \(90.6\%\pm0.7\%\)  & \(97.5\%\pm0.7\%\)  \\
\(K\ge5\)   & \(93.2\%\pm0.6\%\)  & \(88.6\%\pm0.9\%\)  & \(95.7\%\pm0.9\%\)  \\
\(K\ge7\)   & \(90.2\%\pm0.9\%\)  & \(84.7\%\pm0.6\%\)  & \(91.4\%\pm1.3\%\)  \\
\(K\ge10\)  & \(84.5\%\pm0.6\%\)  & \(80.4\%\pm0.9\%\)  & \(82.8\%\pm1.7\%\)  \\
\(K\ge15\)  & \(75.3\%\pm0.5\%\)  & \(75.4\%\pm0.7\%\)  & \(63.5\%\pm1.7\%\)  \\
Majority    & \(74.0\%\pm0.9\%\)  & \(74.4\%\pm0.7\%\)  & \(60.1\%\pm2.0\%\)  \\

\bottomrule
\end{tabular}}
\caption{Cross-model comparison of $K$ correct personas out of 31 in ensemble voting (mean and $\pm$ std across runs)}
\label{tab:correct_k}

\end{table}

\paragraph{Oracle cross-model comparison}
The distribution of correct persona counts across samples is further illustrated in Table~\ref{tab:correct_k} for a cross-model comparison. As expected, the models follow different voting behaviors. For example, while gpt-oss-20b achieves the highest oracle accuracy at relaxed thresholds (up to $K \geq 7$), its performance drops sharply at stricter thresholds and falls well behind the other models at the majority vote level. This suggests that gpt-oss-20b exhibits high per-sample oracle coverage but low inter-persona agreement: individual personas frequently arrive at the correct answer, yet they rarely converge on it collectively.

\begin{table}[htbp]
\centering
\footnotesize 
\setlength{\tabcolsep}{3pt} 
\resizebox{\columnwidth}{!}{
\begin{tabular}{clcccc} 
\toprule
\textbf{Rank} & \textbf{Persona} & \textbf{All} & \textbf{Qwen3} & \textbf{Gemma} & \textbf{gpt-oss} \\
\midrule
-- & \textbf{MAJORITY VOTES ALL} & \textbf{75.8} & \textbf{74.3} & \textbf{74.5} & \textbf{60.8} \\
-- & \textbf{MAJORITY [P] only} & \textbf{76.0} & \textbf{74.2} & \textbf{74.1} & \textbf{60.2} \\
-- & \textbf{MAJORITY [P] (subsamp=13)} & \textbf{75.8} & \textbf{74.1} & \textbf{74.1} & \textbf{60.4} \\
-- & \textbf{MAJORITY [L] only} & \textbf{75.3} & \textbf{73.5} & \textbf{74.5} & \textbf{59.1} \\


\midrule
1 &   [L] Tour Guide & 67.70 & 71.70 & 73.60 & 57.60\\
2 &  [P] Activist & 67.20 & 70.30 & 74.20 & 57.10\\
3 &  [L] Small Business Owner & 67.10 & 70.20 & 74.20 & 56.90\\
4 & [L] Football Player & 67.00 & 69.30 & 74.00 & 57.80\\
5 &  [L] Taxi Driver & 67.00 & 70.90 & 74.40 & 55.70\\
6 &  [P] Storyteller & 67.00 & 69.60 & 72.80 & 58.40\\
7 &  [P] Concise Reader & 66.90 & 68.70 & 73.50 & 58.30\\
8 &  [P] Data Scientist & 66.70 & 69.20 & 73.70 & 57.20\\
9 & [P] Computational Narratologist & 66.70 & 68.60 & 72.50 & 59.00\\
10 & [L] Barkeeper & 66.70 & 69.30 & 74.20 & 56.50\\
11 &   [L] High School Student & 66.60 & 68.20 & 74.50 & 57.10\\
12 &   [L] Fitness Coach & 66.60 & 68.90 & 72.00 & 58.80\\
13 & [P] Literary Critic & 66.60 & 69.10 & 72.00 & 58.50\\
14 & [P] Computer Scientist & 66.50 & 69.40 & 71.80 & 58.30\\
15 &   [L] Construction Worker & 66.40 & 70.10 & 72.40 & 56.60\\
16 &  [L] Nurse & 66.40 & 69.70 & 73.30 & 56.00\\
17 &  [P] Hermeneutics Specialist & 66.20 & 69.90 & 72.40 & 56.10\\
18 &   [P] Journalist & 66.10 & 69.50 & 71.20 & 57.50\\
19 &  [P] Critical Theorist & 66.10 & 70.90 & 69.00 & 58.30\\
20 &   [L] Retiree & 66.10 & 68.20 & 72.90 & 57.00\\
21 & [L] Pirate & 66.00 & 70.30 & 72.40 & 55.40\\
22 &  [L] Electrician & 65.90 & 69.40 & 70.60 & 57.50\\
23 &  [P] Ethicist & 65.80 & 70.40 & 70.90 & 56.10\\
24 & [L] Line Cook & 65.70 & 68.80 & 72.50 & 55.80\\
25 & [P] Psychologist & 65.70 & 67.00 & 71.80 & 58.00\\
26 &  [P] Yellow Press Journalist & 65.60 & 69.60 & 72.30 & 54.90\\
27 & [P] Fact Checker & 65.20 & 68.60 & 69.20 & 57.70\\
28 &   [P] Psychoanalyst & 64.90 & 69.10 & 68.20 & 57.30\\
29 &  [P] Gender Perspective Analyst & 64.40 & 67.30 & 69.70 & 56.10\\
30 &   [P] Postcolonial Critic & 64.30 & 70.20 & 67.00 & 55.70\\
31 &  [P] Feminist Literature Critic & 63.60 & 67.80 & 66.20 & 56.70\\
\bottomrule
\end{tabular}%
}
\caption{Per-Persona Accuracy (\%) Comparison Across Models, sorted by the ``All'' column. [L] is short for Lay people, [P] for Practitioners.}
\label{tab:persona_accuracy}
\end{table}



Table~\ref{tab:persona_accuracy} showcases the results across these configurations. Model-specific and cross-model majority votes consistently outperform individual persona predictions, with one exception: the ``High School Student'' persona matches the model-specific majority vote accuracy on Gemma~27B-it, though it is outperformed by the cross-model majority vote. The performance also differs across persona categories: [L]~personas dominate the top~5 individual results while [P]~personas occupy the bottom~5. In contrast, majority voting gives [P] an advantage over [L]. Since the category sizes are imbalanced (18~[P] vs.\ 13~[L]), we also perform a subsampling analysis where we randomly choose 13~[P] personas over 500 iterations, achieving a mean accuracy of 75.8\% ($\sigma = 0.3\%$, 95\% CI [75.2\%, 76.4\%]), comparable to the full [P] set (76.0\%).

\paragraph{Further Results}

On the test set, the simple majority vote achieved 0.7025 accuracy. An optimized $k{=}8$ persona subset, selected via exhaustive search with 5-fold cross-validation, reached $82.0\%\pm4.8\%$ on dev but only 0.705 on test, which is nearly identical to the majority vote (details in Appendix~\ref{sec:appendix_exhaustive_search}). The 11.5 percentage point dev-test gap (vs.\ 5.6 for the full ensemble) suggests overfitting to the small dev set (200 items), indicating that broad ensemble aggregation is more robust than targeted subset optimization for this task.

\section{Error Analysis}

To further investigate the performance split between [P] and [L] found in Table \ref{tab:persona_accuracy} and to specifically analyze why the ``Gender Perspective Analyst'' and ``Feminist Literary Critic'' (grouped together as [$P_G$]) personas particularly underperformed, we run a series of error and statistical analyses. We choose these personas because of their low performance, their similarity to each other (compared to the other personas) and because feminist literary criticism is among the most established approaches in literary studies \cite{plain2007history, cooke2020new}. Given the rich theoretical and methodological foundation, the underperformance of these two personas is unexpected and warrants closer investigation if gender-analytical framing might interfere with narrative similarity judgments.

\paragraph{Error Diversity Analysis}

A consistency analysis across persona ensembles shows that [$P_G$] are among the least consistent voters, while ensemble disagreement proxies item difficulty, with high-agreement items corresponding to clearer narrative similarity cases (see Appendix \ref{app:consistency}).
Table \ref{tab:persona_accuracy} reveals that [P] personas generally perform worse individually yet achieve a higher majority vote accuracy. To investigate this behavior, we perform a pairwise error correlation analysis considering individual LLM persona outputs and group performance. Following \cite{Kuncheva2003}, we compute pairwise error correlation and double-fault rate across personas within each category. The results in Table \ref{tab:error_analysis} point at one key difference between the groups: Members of [L] vote more unanimously, while votes from members of [P] with more concretely formulated perspectives vote more distinctly (further details in Appendix \ref{sec:error_analysis_2}).

\begin{table}[t]
\centering
\small
\begin{tabular}{p{2.5cm} p{1cm} p{1cm} p{2.5cm}}
\toprule
\textbf{Metric} & \textbf{[P]} & \textbf{[L]} & \textbf{Conclusion} \\
\midrule
Individual accuracy & 71.0\% & 71.7\% & Higher accuracy for [L] \\
\addlinespace
Error correlation ($r$) & 0.388 & 0.461 & Errors are 19\% less correlated for [P] \\
\addlinespace
\begin{tabular}[c]{@{}l@{}}Double-fault \\ $P(\text{both wrong})$\end{tabular}
& 0.164 & 0.173 & [L] fail together more often \\
\bottomrule
\end{tabular}
\caption{Error diversity analysis: comparison of [P] and [L] groups.
Individual metrics are averaged across personas within each category;
diversity metrics are averaged across all pairwise persona combinations
within each category. Following \cite{Kuncheva2003}.}
\label{tab:error_analysis}
\end{table}

\paragraph{Gender/Feminist vocabulary correlation}
\label{sec:gender_correlation_main}
\begin{table*}[t]
\centering
\small
\begin{tabular}{@{}llrrrrr@{}}
\toprule
\textbf{Direction} & \textbf{Term} & \textbf{$r_{pb}$} & \textbf{$p_{\text{FDR}}$} & \textbf{Acc (present)} & \textbf{Acc (absent)} & \textbf{$N$} \\
\midrule
\multicolumn{7}{l}{\textit{Negative correlations (gender vocabulary associated with lower accuracy)}} \\
\midrule
$\downarrow$ & \texttt{female}              & $-$0.046 & $<$0.001 & 56.5\% & 66.6\% & 4{,}575 \\
$\downarrow$ & \texttt{gender roles}        & $-$0.039 & $<$0.001 & 42.0\% & 66.2\% & 571 \\
$\downarrow$ & \texttt{objectification}     & $-$0.032 & $<$0.001 & 13.6\% & 66.1\% & 81 \\
$\downarrow$ & \texttt{female protagonist}  & $-$0.030 & $<$0.001 & 54.9\% & 66.3\% & 1{,}516 \\
$\downarrow$ & \texttt{sexual violence}     & $-$0.029 & $<$0.001 & 34.9\% & 66.2\% & 186 \\
$\downarrow$ & \texttt{female character}    & $-$0.027 & $<$0.001 & 47.4\% & 66.2\% & 430 \\
$\downarrow$ & \texttt{experiences husband} & $-$0.024 & $<$0.001 & 0.0\%  & 66.1\% & 28 \\
$\downarrow$ & \texttt{gender}              & $-$0.022 & $<$0.001 & 56.3\% & 66.2\% & 1{,}102 \\
$\downarrow$ & \texttt{traditional gender}  & $-$0.022 & $<$0.001 & 24.6\% & 66.1\% & 61 \\
$\downarrow$ & \texttt{patriarchal}         & $-$0.021 & $<$0.001 & 49.3\% & 66.1\% & 337 \\
\midrule
\multicolumn{7}{l}{\textit{Positive correlations (gender vocabulary associated with higher accuracy)}} \\
\midrule
$\uparrow$ & \texttt{agency}                & $+$0.019 & $<$0.001 & 71.7\% & 65.9\% & 2{,}547 \\
$\uparrow$ & \texttt{fulfillment outside}   & $+$0.012 & 0.005    & 96.8\% & 66.1\% & 31 \\
$\uparrow$ & \texttt{individuals babies}    & $+$0.012 & 0.005    & 100.0\% & 66.1\% & 25 \\
$\uparrow$ & \texttt{physical danger}       & $+$0.011 & 0.008    & 89.6\% & 66.1\% & 48 \\
$\uparrow$ & \texttt{women within}          & $+$0.011 & 0.009    & 87.5\% & 66.1\% & 56 \\
\bottomrule
\end{tabular}
\caption{Point-biserial correlations between presence of gender/feminist-distinctive vocabulary and voting accuracy for \textit{Other Expert} personas (combined across all three models, all output fields). Terms were identified via lift-based analysis ($\text{lift} \geq 5.0$) from the two gender/feminist persona outputs. Correlations are FDR-corrected (Benjamini--Hochberg, $\alpha = 0.05$). $N$ = number of responses containing the term. Full results for all persona groups and per-model breakdowns are provided in Appendix~\ref{app:gender_correlation}.}
\label{tab:gender_term_correlation_main}
\end{table*}


To investigate whether gender-focused interpretation affects accuracy, we collected vocabulary distinctive to [$P_G$] from LLM outputs (including explanations, themes and key points) using a lift-based approach: for each term, we computed the ratio of its relative frequency in [$P_G$] outputs to its relative frequency across all persona outputs, retaining terms with lift $\geq$ 5.0 as characteristic of gender-focused interpretation. We then measured point-biserial correlations between the presence of these terms and voting correctness across all persona groups (Table~\ref{tab:gender_term_correlation_main}). The results reveal a predominantly negative association: when other members of [P] use gender-distinctive terms, their accuracy drops compared to responses where these terms are absent. While statistically significant after FDR correction, these correlations are small in magnitude ($|r_{pb}| \leq 0.046$), reflecting the large sample sizes involved; the pattern is therefore best understood as a consistent directional tendency rather than a strong predictive signal. It holds across all three models and extends to Lay personas as well as [$P_G$] themselves (Appendix~\ref{app:gender_correlation}).

\section{Discussion}
\label{sec:discussion}

Our results highlight a specific relation between individual persona accuracy and collective decision quality. Table~\ref{tab:error_analysis} can be read as an argument for interpretive diversity: while [P] personas individually underperform [L] personas, their more distinctly formulated perspectives produce less correlated errors (19\% lower pairwise error correlation), which in turn yields a larger ensemble gain under majority voting (75.3\% vs 76.0\%). This aligns with a widely held intuition in ensemble learning that diversity of errors can compensate for lower individual accuracy under majority voting \cite{Kuncheva2003}, although the relationship between diversity measures and ensemble accuracy remains complex \citep{tekin-etal-2024-llm}. 
While all personas exceed 50\% accuracy (satisfying CJT's competence condition), the independence assumption is violated: Table~\ref{tab:error_analysis} confirms substantial pairwise error correlations ($r{=}.388$ for [P], $r{=}.461$ for [L]), which can limit majority vote gains in LLM-based ensembles \cite{lefort2024}. Nevertheless, cross-model majority voting yields clear improvements over individual performance (75.8\% vs.\ 67.7\% maximum), suggesting that the combination of distinct persona framings and separate model families introduces sufficient diversity for meaningful gains under weakened independence.

The near-identical test performance of simple majority voting and the optimized $k{=}8$ ensemble (0.7025 vs.\ 0.705) reinforces this point from a practical perspective. Despite achieving 82.0\% on the dev set through exhaustive subset search, the optimized ensemble offered no meaningful advantage on unseen data. This suggests that for interpretive tasks where item-level difficulty and annotation variability are high, aggregating a large and diverse set of perspectives is more robust than selecting a small optimized subset. The finding also aligns with the broader argument of the paper: in settings where multiperspectivity is a feature rather than a source of noise, the diversity of interpretation may matter more than precision of selection.


The negative correlation between gender-analytical vocabulary and accuracy admits two complementary readings. Gender-focused framing may attend to thematic aspects that, while interpretively plausible, are not predictive of the similarity judgments encoded in the benchmark. Alternatively, these terms may reflect valid interpretations unrepresented in the ground truth. Either way, if valid interpretive perspectives can be penalized by standard accuracy metrics, the question of how benchmarks should account for multiperspectivity becomes pressing as NLP tasks move deeper into interpretive territory \citep{Kommers2025}.

\section*{Limitations}

Several limitations of this study should be acknowledged. 
First, our experiments investigate majority voting and ensemble optimization but do not evaluate multi-agent debate (MAD) systems, in which personas iteratively discuss and revise their judgments. Our results therefore do not establish that simple voting outperforms deliberative approaches; rather, they characterize the behavior of independent, non-interacting ensembles. Evaluating whether structured interaction between personas yields additional gains remains an important direction for future work.
Second, while our statistical analyses reveal significant correlations between vocabulary usage and accuracy, correlation does not equal causation. The presence of gender-analytical terms may co-occur with particular narrative properties of the input texts rather than causally driving incorrect predictions. Our discussion of these findings is itself interpretive, and alternative explanations remain plausible.
Third, no conclusions should be drawn about the demographic or professional backgrounds of the dataset's annotators from our persona-level results. The observation that lay personas like the Tour Guide outperform the Feminist Literary Critic does not imply that the annotator pool was composed of individuals resembling the former more than the latter; it reflects properties of the LLM role-playing behavior and the specific benchmark, not of human annotation practices.
Finally, our experiments are conducted on a single benchmark (SemEval-2026 Task~4) with a development set of 200 items. While we observe consistent patterns across three models, the generalizability of our findings to other narrative similarity datasets or related interpretive tasks remains to be established.


\bibliography{custom}
\clearpage
\appendix


\section*{Appendix}

\section{Consistency Analysis}
\label{app:consistency}

\begin{table}[h]
\centering

\small
\resizebox{\columnwidth}{!}{%
\begin{tabular}{lcccc}
\toprule
\textbf{Persona} & \textbf{All} & \textbf{Qwen3} & \textbf{Gemma} & \textbf{gpt-oss} \\
\midrule
Yellow Press Journalist      & 83.3 & 83.9 & 91.8 & 74.2 \\
Fitness Coach                & 83.1 & 83.7 & 93.0 & 72.7 \\
Computational Narratologist  & 83.1 & 83.3 & 92.1 & 73.7 \\
Psychologist                 & 82.9 & 83.5 & 93.1 & 72.2 \\
Electrician                  & 82.8 & 83.4 & 91.8 & 73.2 \\
Ethicist                     & 82.8 & 83.2 & 93.1 & 72.1 \\
Critical Theorist            & 82.7 & 84.0 & 92.3 & 71.8 \\
Small Business Owner         & 82.7 & 83.6 & 92.5 & 72.0 \\
High School Student          & 82.6 & 83.3 & 91.5 & 73.0 \\
Barkeeper                    & 82.5 & 83.3 & 91.0 & 73.3 \\
Tour Guide                   & 82.5 & 83.8 & 91.8 & 72.0 \\
Activist                     & 82.5 & 83.0 & 93.0 & 71.6 \\
Postcolonial Critic          & 82.5 & 83.7 & 92.2 & 71.5 \\
Nurse                        & 82.4 & 82.3 & 92.5 & 72.2 \\
Concise Reader               & 82.3 & 83.2 & 91.2 & 72.5 \\
Construction Worker          & 82.3 & 83.1 & 91.2 & 72.7 \\
Retiree                      & 82.3 & 83.3 & 90.9 & 72.6 \\
Line Cook                    & 82.2 & 82.9 & 91.7 & 72.1 \\
Psychoanalyst                & 82.2 & 84.0 & 91.9 & 70.7 \\
Taxi Driver                  & 82.1 & 81.8 & 91.2 & 73.2 \\
Data Scientist               & 82.0 & 80.9 & 91.0 & 74.0 \\
Hermeneutics Specialist      & 82.0 & 84.4 & 90.5 & 71.0 \\
Journalist                   & 82.0 & 82.0 & 90.7 & 73.3 \\
Literary Critic              & 81.7 & 81.8 & 91.5 & 71.9 \\
Computer Scientist           & 81.6 & 82.2 & 90.7 & 71.9 \\
Football Player              & 81.6 & 81.4 & 91.8 & 71.6 \\
Pirate                       & 81.5 & 80.2 & 92.5 & 71.8 \\
Storyteller                  & 81.4 & 81.3 & 91.2 & 71.8 \\
Feminist Literature Critic   & 81.0 & 81.1 & 90.7 & 71.4 \\
Gender Perspective Analyst   & 80.9 & 81.1 & 90.2 & 71.3 \\
Fact Checker                 & 80.6 & 80.7 & 89.1 & 72.2 \\
\midrule
\textbf{Average}             & \textbf{82.2} & \textbf{82.7} & \textbf{91.6} & \textbf{72.3} \\
\bottomrule
\end{tabular}

}
\caption{Voting consistency (\%) per persona and model. Consistency is defined as the fraction of runs matching the modal (most common) vote for each item, averaged across all 200 items. Each persona was evaluated 10 times per item; a consistency of 100\% means the persona always gave the same answer across all runs, while 50\% indicates a maximally split vote. Personas are sorted by combined consistency (descending).}
\label{tab:consistency}
\end{table}
To measure prediction certainty as a potential error indicator and to see if it correlates with accuracy, we opt for sampling consistency over verbalized (self-reported) confidence scores: we run each persona 10 times at temperature $t{=}1$ and measure the proportion of runs that yield the same predicted label. This follows works showing that consistency across stochastic samples provides a more reliable uncertainty signal than verbalized confidence, which tends to be poorly calibrated and systematically overconfident in instruction-tuned LLMs \cite{xiong2024can,manakul2023selfcheckgpt}. 

To empirically validate this choice, we compared both signals as accuracy predictors at all three analysis levels reported below. Across all levels and models, sampling consistency was the substantially stronger predictor: at the item level, consistency correlated with accuracy at $r = +0.53$ (combined), while verbalized confidence showed near-zero correlation ($r = +0.10$). At the within-persona level, consistency yielded significant positive correlations for all
31 personas across every model (mean $\bar{r} = +0.42$ combined), whereas verbalized confidence reached significance for only 21 of 31 personas with a mean $\bar{r}$ of just $+0.07$. These results confirm that, consistent with prior findings, sampling consistency is the more reliable uncertainty proxy for this setting.

As shown in table~\ref{tab:consistency}, [$P_G$] predictions are indeed among the least consistent. To check if more consistent personas tend to be more accurate, we analyzed correlations at three levels (all $p$-values were FDR-corrected using the Benjamini-Hochberg procedure):

\begin{itemize}
    \item \textbf{Across-persona level:} The Pearson correlation between mean consistency and mean accuracy was non-significant for individual models (Qwen-14B: $r = +0.21$, $p_{\mathrm{adj}} = .34$; Gemma-27B: $r = +0.16$, $p_{\mathrm{adj}} = .43$; OSS-20B: $r = -0.00$, $p_{\mathrm{adj}} = .99$), but reached significance in the combined analysis ($r = +0.53$, $p_{\mathrm{adj}} = .004$).
    \item \textbf{Within-persona level:} Per item, all 31 personas showed significant positive correlations ($p_{\mathrm{adj}} < .05$) across every model, with mean $\bar{r}$ values of $+0.43$ (Qwen-14B), $+0.30$ (Gemma-27B), $+0.27$ (OSS-20B), and $+0.55$ (combined).
    \item \textbf{Item level:} Averaging consistency and accuracy across all personas per item yielded Pearson $r$ values of $+0.52$ (Qwen-14B), $+0.44$ (Gemma-27B), $+0.37$ (OSS-20B), and $+0.61$ combined (all $p < .001$; Spearman $\rho = +0.75$ combined).
\end{itemize}

Together, these results indicate that while the personas themselves have little effect on the consistency-accuracy correlation, item difficulty is a dominant factor: items on which personas agree tend to be answered correctly, while items with split votes are disproportionately likely to be wrong. Thus, ensemble disagreement can serve as a proxy for item difficulty and, by extension, expected accuracy.

\section{Ensemble Agreement as Inter-Annotator Agreement}
To examine whether interpretive framing meaningfully influences similarity judgments, we treat each persona as an annotator: for each item, we collapse the 10 stochastic runs into a single label via majority vote, then compute Krippendorff's $\alpha$ (nominal) over the resulting annotation matrix. Per individual model, agreement is substantial for Qwen~3-14B ($\alpha = 0.73$) and Gemma~27B ($\alpha = 0.72$) but only moderate for gpt-oss-20B ($\alpha = 0.49$). When all 93 persona\,\texttimes\,model combinations are treated as independent annotators, agreement drops to $\alpha = 0.42$ (moderate). Notably, [P]~personas consistently show lower agreement than [L]~personas (combined $\alpha = 0.40$ vs.\ $0.45$; per-model differences of 6--13~percentage points), mirroring the lower pairwise error correlations reported in Table~\ref{tab:error_analysis}.

These values can be situated relative to the benchmark's own inter-annotator agreement: \citet{hatzel2026} report $\alpha = 0.33$ for pairwise human annotations, noting that the dataset was filtered to retain only difficult cases. That the persona ensemble achieves comparable or higher agreement than human annotators ($\alpha = 0.42$ combined, up to $0.73$ per model) while still falling short of near-perfect consensus reinforces that narrative similarity judgments involve genuine interpretive variability, which is a property of the task itself, not merely an artifact of LLM generation noise.

\clearpage
\section{LLM Personas}
\label{sec:llm_persone}

The full list of LLM personas is documented in the following:



\onecolumn

\small
\begin{longtable}{p{3.5cm} p{2.5cm} p{8.5cm}}

\toprule
\textbf{LLM Persona} & \textbf{Role} & \textbf{System Prompt} \\
\midrule
\endfirsthead

\toprule
\textbf{LLM Persona} & \textbf{Role} & \textbf{System Prompt} \\
\midrule
\endhead

\midrule
\multicolumn{3}{r}{\emph{Continued on next page}} \\
\bottomrule
\endfoot

\bottomrule
\endlastfoot

Literary Critic & Practitioner & You are a Literary Critic analyzing narrative similarity. \\
Computational Narratologist & Practitioner & You are a Computational Narratologist. You excel at structural analysis of text and the structured extraction of narrative features, such as actors, relations, events, topics, themes, sentiments etc. \\
Psychologist & Practitioner & You are a Psychologist analyzing emotional and motivational similarity. \\
Psychoanalyst & Practitioner & You are a Psychoanalyst analyzing the text from the perspective of engaging with the unconscious and how it relates to society, focusing on authorial intent, symbolism and conflicts. \\
Storyteller & Practitioner & You are a Storyteller analyzing narrative flow and engagement. \\
Computer Scientist & Practitioner & You are a Computer Scientist analyzing structural consistency. You don't care much for literature, but you are able to talk about text. \\
Data Scientist & Practitioner & You are a Data Scientist checking consistency in the data. \\
Ethicist & Practitioner & You are an Ethicist Critic analyzing normative implications. \\
Critical Theorist & Practitioner & You are a Critical Theorist analyzing the texts in regard to its function within late-capitalist society. \\
Postcolonial Critic & Practitioner & You are a Postcolonial Critic examining the texts in terms of global power dynamics. \\
Feminist Literary Critic & Practitioner & You are a Feminist Literary Critic examining gender-related dynamics within texts. \\
Gender Perspective Analyst & Practitioner & You are a Gender Perspective Analyst examining gender-related dynamics. \\
Fact Checker & Practitioner & You are a Fact-Checker verifying claims. \\
Concise Reader & Practitioner & You are a Concise Reader giving very short summaries. \\
Hermeneutics Specialist & Practitioner & You are a specialist in hermeneutics in the tradition of Paul Ricœur. \\
Activist & Practitioner & You fight for a better world and are an expert in political communication. \\
Journalist & Practitioner & You are a Journalist able to capture the narrative of any given text quickly. \\
Yellow Press Journalist & Practitioner & You are a Yellow Press Journalist. You are a specialist in writing texts which get everyone's attention. \\
\midrule
High School Student & Lay & You are an 8th grader. You are not the best student, but you are doing alright. \\
Football Player & Lay & You are a professional football player and not sure why you got this task, but provide your perspective anyway. \\
Pirate & Lay & You are a pirate from the 16th century. Please provide your perspective. \\
Barkeeper & Lay & You are a barkeeper and spend a lot of time talking to many people from all walks of life. \\
Construction Worker & Lay & You are a veteran construction worker. You value things that are built on a solid foundation. \\
Nurse & Lay & You are a registered nurse. You are empathetic but very grounded in reality. \\
Taxi Driver & Lay & You are a taxi driver who has driven thousands of miles and heard a million stories. \\
Line Cook & Lay & You are a line cook in a busy diner. You have a blunt, no-nonsense attitude. \\
Fitness Coach & Lay & You are a personal trainer and fitness coach. You focus on action, momentum, and results. \\
Electrician & Lay & You are an electrician. You look for the 'current' in a story—how things are connected. \\
Small Business Owner & Lay & You run a local hardware store. You are practical and budget-conscious. \\
Tour Guide & Lay & You are a local city tour guide. You explain complex histories in engaging ways. \\
Retiree & Lay & You are a retired office manager with a straightforward, common-sense approach. \\
\end{longtable}
\label{tab:personas_full}
\twocolumn

\section{Exhaustive Search Results for Voting Ensemble Optimization}
\label{sec:appendix_exhaustive_search}

\begin{table}[t]
\centering
\begin{tabular}{c c c}
\hline
\textbf{Ensemble Size} & \textbf{Accuracy (\%)} & \textbf{Std (\%)} \\
\hline
1  & 77.50 & $\pm$7.60 \\
7  & 81.50 & $\pm$5.40 \\
8  & 82.00 & $\pm$4.80 \\
9  & 81.50 & $\pm$7.30 \\
15 & 81.00 & $\pm$5.60 \\
\hline
\end{tabular}
\caption{Exhaustive Search results: Ensemble size vs. accuracy (mean $\pm$ std).}
\label{tab:exhaustive_search}
\end{table}

\normalsize
To find optimal ensemble sizes consisting of different voting patterns we performed exhaustive search over all $\binom{n}{k}$ possible k-member subsets from the top-n=31 candidates. Each candidate ensemble was evaluated using 5-fold cross-validation with simple majority voting. We selected the ensemble size by maximizing accuracy. When the number of combinations exceeded 50,000, we approximated exhaustive search via random sampling. Table \ref{tab:exhaustive_search} shows an evaluation of different ensemble sizes in relation to accuracy.

\begin{table*}[h]
\centering
\caption{Error diversity analysis comparing [P] and [L] persona categories. 
}
\label{tab:error_analysis_2}
\small
\setlength{\tabcolsep}{4pt}
\begin{tabular}{@{}l cc c cc c cc c cc@{}}
\toprule
& \multicolumn{2}{c}{\textbf{Combined}} 
&& \multicolumn{2}{c}{\textbf{Qwen3-14B}} 
&& \multicolumn{2}{c}{\textbf{Gemma-3-27B}} 
&& \multicolumn{2}{c}{\textbf{GPT-OSS-20B}} \\
\cmidrule{2-3} \cmidrule{5-6} \cmidrule{8-9} \cmidrule{11-12}
& [P] & [L] && [P] & [L] && [P] & [L] && [P] & [L] \\
\midrule
\multicolumn{12}{@{}l}{\textit{Panel A: Individual \& Ensemble Performance}} \\[3pt]
Number of personas          & 18    & 13    && 18    & 13    && 18    & 13    && 18    & 13    \\
Mean individual acc.\ (\%)  & 71.0  & 71.7  && 69.2  & 69.6  && 71.0  & 73.1  && 57.3  & 56.8  \\
Accuracy std (\%)           & 1.0   & 0.6   && ---   & ---   && ---   & ---   && ---   & ---   \\
Accuracy range (\%)         & 3.6   & 1.9   && ---   & ---   && ---   & ---   && ---   & ---   \\
Majority vote acc.\ (\%)    & 76.0  & 75.3  && 74.2  & 73.5  && 74.1  & 74.5  && 60.2  & 59.1  \\
\addlinespace[2pt]
\quad Ensemble gain$^\dagger$ (\%)  & +5.0  & +3.6  && +5.0  & +3.9  && +3.1  & +1.4  && +2.9  & +2.3  \\
\midrule
\multicolumn{12}{@{}l}{\textit{Panel B: Pairwise Error Diversity (lower = more diverse)}} \\[3pt]
Vote agreement              & .748  & .782  && ---   & ---   && ---   & ---   && ---   & ---   \\
Pearson $r$ (correctness)   & .388  & .461  && .366  & .406  && .516  & .620  && .211  & .236  \\
Cohen's $\kappa$            & .387  & .461  && .366  & .406  && .515  & .620  && .211  & .236  \\
Double-fault rate           & .164  & .173  && .173  & .177  && .190  & .194  && .234  & .244  \\
$P(\text{both} \mid \geq\!1)$  & .395  & .443  && ---   & ---   && ---   & ---   && ---   & ---   \\
\midrule
\multicolumn{12}{@{}l}{\textit{Panel C: Expert--Lay Differences ($\Delta = \text{Expert} - \text{Lay}$)}} \\[3pt]
$\Delta$ Individual acc.\ (\%)     & $-$0.7 & && $-$0.4 & && $-$2.1 & && $+$0.5 &  \\
$\Delta$ Majority vote acc.\ (\%)  & $+$0.7 & && $+$0.7 & && $-$0.4 & && $+$1.1 &  \\
$\Delta$ Pearson $r$               & $-$.074 & && $-$.040 & && $-$.104 & && $-$.025 &  \\
$\Delta$ Cohen's $\kappa$          & $-$.074 & && $-$.040 & && $-$.105 & && $-$.025 &  \\
$\Delta$ Double-fault rate         & $-$.010 & && $-$.005 & && $-$.004 & && $-$.010 &  \\
\bottomrule
\end{tabular}

\vspace{4pt}
{\footnotesize 
$^\dagger$Ensemble gain = majority vote accuracy $-$ mean individual accuracy. \\
Combined results aggregate persona votes across all three models per evaluation case.\\
Entries marked --- were computed only for the combined setting.
}
\end{table*}

The resulting selection consisted of ``Computer Scientist'', ``Critical Theorist'', ``Electrician'', ``Fitness Coach'', ``Hermeneutics Specialist`` from Qwen3 14B and ``Data Scientist'', ``Tour Guide'', ``Yellow Press Journalist'' from Gemma 3 27B-it while notably no personas from gpt-oss-20B were included by this selection strategy.

\section{Extended Error Diversity Analysis}
\label{sec:error_analysis_2}

Table \ref{tab:error_analysis_2} presents our extended error diversity analysis which focuses on collected metrics per model. Individual metrics are averaged across personas within each category; 
diversity metrics are averaged across all pairwise persona combinations within each category.
All diversity metrics are consistently lower for [P] across settings, 
indicating more diverse error patterns despite lower individual accuracy.

\section{Gender/Feminist Vocabulary Correlation Analysis}
\label{app:gender_correlation}

\subsection{Method}
\label{app:gender_method}

We identified vocabulary distinctive to [$P_G$] relative to the remaining 16 personas in [P] and 13 in [L] using a lift-based approach. For each term $t$ (unigrams and bigrams, extracted via NLTK \citep{bird2009natural} after lowercasing, stopword removal, and filtering to alphabetic tokens of length $>1$), we computed $\text{lift}(t) = (f_{\text{gender}}(t) + \epsilon) / (f_{\text{other}}(t) + \epsilon)$ with $\epsilon = 10^{-8}$. A term was classified as \textit{gender/feminist-distinctive} if $\text{lift}(t) \geq 5.0$ and it occurred at least 10 times in gender persona outputs. We analyzed four output fields per persona: \textit{themes}, \textit{key points}, \textit{explanation}, and \textit{all} (their concatenation).

This procedure identified 229 (Qwen3-14B), 935 (Gemma-27B), and 216 (GPT-OSS-20B) distinctive terms on the \textit{all} field, with 1{,}223 in the combined analysis. Representative terms range from demographic markers (\textit{female protagonist}, \textit{male lead}) and analytical concepts (\textit{gender roles}, \textit{patriarchal}, \textit{agency}) to relational terms (\textit{caregiving}, \textit{male bonding}) and critical vocabulary (\textit{objectification}, \textit{gendered}).

For each distinctive term $t$ and persona group [$P_G$], [$P \setminus G$], [L], we computed the point-biserial correlation $r_{pb}(t, G) = \text{corr}(\mathbf{1}[t \in \text{response}],\; \mathbf{1}[\text{vote} = \text{GT}])$, excluding ties and requiring $\geq 5$ occurrences per group. Multiple testing was controlled via Benjamini--Hochberg FDR correction \citep{benjamini1995controlling} at $\alpha = 0.05$, applied independently within each persona group $\times$ output field combination.%

\subsection{Summary of Significant Correlations}
\label{app:gender_corr_summary}

Table~\ref{tab:gender_corr_summary} provides an overview of the number and direction of significant correlations across all analysis conditions.

\begin{table*}[h]
\centering
\small
\begin{tabular}{@{}ll rrr rrr rrr@{}}
\toprule
& & \multicolumn{3}{c}{\textbf{Gender/Fem.\ (self)}} & \multicolumn{3}{c}{\textbf{Other Experts}} & \multicolumn{3}{c}{\textbf{Lay Personas}} \\
\cmidrule(lr){3-5} \cmidrule(lr){6-8} \cmidrule(lr){9-11}
\textbf{Model} & \textbf{Field} & Sig. & $+$ & $-$ & Sig. & $+$ & $-$ & Sig. & $+$ & $-$ \\
\midrule
\multirow{4}{*}{Qwen3-14B}
  & themes      &  4 & 0 &  4 & 11 & 1 & 10 & 10 & 0 & 10 \\
  & key\_points &  5 & 0 &  5 & 12 & 0 & 12 & 11 & 1 & 10 \\
  & explanation &  6 & 0 &  6 & 22 & 0 & 22 & 19 & 0 & 19 \\
  & all         & 17 & 0 & 17 & 37 & 1 & 36 & 25 & 1 & 24 \\
\midrule
\multirow{4}{*}{Gemma-27B}
  & themes      & 96 & 19 & 77 & 79 & 15 & 64 & 40 &  4 & 36 \\
  & key\_points & 30 &  1 & 29 & 43 & 10 & 33 & 16 &  0 & 16 \\
  & explanation & 41 &  1 & 40 & 41 &  4 & 37 & 23 &  1 & 22 \\
  & all         &149 & 15 &134 &137 & 25 &112 & 82 &  4 & 78 \\
\midrule
\multirow{4}{*}{GPT-OSS-20B}
  & themes      &  3 & 0 &  3 &  0 & 0 &  0 &  0 & 0 &  0 \\
  & key\_points &  5 & 3 &  2 &  6 & 2 &  4 &  4 & 3 &  1 \\
  & explanation &  0 & 0 &  0 &  0 & 0 &  0 &  0 & 0 &  0 \\
  & all         &  1 & 0 &  1 &  7 & 5 &  2 &  3 & 2 &  1 \\
\midrule
\multirow{4}{*}{Combined}
  & themes      & 99 & 30 & 69 & 75 & 21 & 54 & 48 & 12 & 36 \\
  & key\_points & 19 &  4 & 15 & 35 &  5 & 30 & 19 &  1 & 18 \\
  & explanation & 36 &  2 & 34 & 56 & 12 & 44 & 23 &  0 & 23 \\
  & all         &132 & 27 &105 &119 & 26 & 93 & 63 &  3 & 60 \\
\bottomrule
\end{tabular}
\caption{Number of significant correlations (FDR-corrected, $p < 0.05$) between gender/feminist-distinctive term presence and voting accuracy. \textit{Sig.}~= total significant; $+$~= positive (term presence associated with higher accuracy); $-$~= negative (term presence associated with lower accuracy).}
\label{tab:gender_corr_summary}
\end{table*}

Across all conditions, negative correlations substantially outnumber positive ones. The pattern is consistent across persona groups: gender/feminist personas themselves, other experts, and lay personas all show predominantly negative associations between gender-distinctive vocabulary and accuracy. Gemma-27B yields the most significant correlations, consistent with its larger distinctive term set; GPT-OSS-20B shows very few, suggesting less differentiated gender-related vocabulary.

\subsection{Top Correlations by Persona Group}
\label{app:gender_corr_full}

Table~\ref{tab:corr_merged_all} presents the strongest correlations for each persona group in the combined (all models, \textit{all} fields) analysis.

\begin{table*}[h]
\centering
\small
\begin{tabular}{@{}ll rlrrrr@{}}
\toprule
\textbf{Group} & \textbf{Dir.} & & \textbf{Term} & $r_{pb}$ & \textbf{Acc (+)} & \textbf{Acc ($-$)} & $N$ \\
\midrule
\multirow{8}{*}{\rotatebox[origin=c]{90}{\parbox{2.2cm}{\centering Gender/Fem.\\(self)}}}
  & \multirow{3}{*}{$\uparrow$}
    & 1 & \texttt{male friendship}       & $+$.043 & 91.3\% & 63.9\% & 69 \\
  & & 2 & \texttt{male relationships}    & $+$.036 & 82.5\% & 63.9\% & 103 \\
  & & 3 & \texttt{male bonding}          & $+$.036 & 83.5\% & 63.9\% & 91 \\
\cmidrule(l){2-8}
  & \multirow{5}{*}{$\downarrow$}
    & 1 & \texttt{male character}        & $-$.058 & 45.7\% & 64.4\% & 269 \\
  & & 2 & \texttt{female}                & $-$.057 & 59.6\% & 65.7\% & 3{,}312 \\
  & & 3 & \texttt{affairs political}     & $-$.050 & 0.0\%  & 64.1\% & 17 \\
  & & 4 & \texttt{experiences husband}   & $-$.049 & 0.0\%  & 64.1\% & 16 \\
  & & 5 & \texttt{closeted identity}     & $-$.047 & 0.0\%  & 64.1\% & 15 \\
\midrule
\multirow{8}{*}{\rotatebox[origin=c]{90}{\parbox{2.2cm}{\centering Other\\Experts}}}
  & \multirow{3}{*}{$\uparrow$}
    & 1 & \texttt{agency}                & $+$.019 & 71.7\% & 65.9\% & 2{,}547 \\
  & & 2 & \texttt{fulfillment outside}   & $+$.012 & 96.8\% & 66.1\% & 31 \\
  & & 3 & \texttt{individuals babies}    & $+$.012 & 100.0\% & 66.1\% & 25 \\
\cmidrule(l){2-8}
  & \multirow{5}{*}{$\downarrow$}
    & 1 & \texttt{female}                & $-$.046 & 56.5\% & 66.6\% & 4{,}575 \\
  & & 2 & \texttt{gender roles}          & $-$.039 & 42.0\% & 66.2\% & 571 \\
  & & 3 & \texttt{objectification}       & $-$.032 & 13.6\% & 66.1\% & 81 \\
  & & 4 & \texttt{female protagonist}    & $-$.030 & 54.9\% & 66.3\% & 1{,}516 \\
  & & 5 & \texttt{sexual violence}       & $-$.029 & 34.9\% & 66.2\% & 186 \\
\midrule
\multirow{8}{*}{\rotatebox[origin=c]{90}{\parbox{2.2cm}{\centering Lay\\Personas}}}
  & \multirow{3}{*}{$\uparrow$}
    & 1 & \texttt{agency}                & $+$.014 & 72.2\% & 66.5\% & 1{,}016 \\
  & & 2 & \texttt{protagonist distracted} & $+$.012 & 100.0\% & 66.5\% & 24 \\
  & & 3 & \texttt{woman ultimately}      & $+$.010 & 90.6\% & 66.5\% & 32 \\
\cmidrule(l){2-8}
  & \multirow{5}{*}{$\downarrow$}
    & 1 & \texttt{female}                & $-$.041 & 57.4\% & 67.0\% & 3{,}320 \\
  & & 2 & \texttt{sexual violence}       & $-$.039 & 20.5\% & 66.6\% & 122 \\
  & & 3 & \texttt{female protagonist}    & $-$.034 & 53.7\% & 66.8\% & 1{,}182 \\
  & & 4 & \texttt{gender roles}          & $-$.032 & 37.2\% & 66.6\% & 207 \\
  & & 5 & \texttt{objectification}       & $-$.027 & 6.1\%  & 66.6\% & 33 \\
\bottomrule
\end{tabular}
\caption{Top significant correlations (FDR-corrected, $p < 0.05$) between gender/feminist vocabulary and voting accuracy, combined across all models, \textit{all} fields. Top 3 positive ($\uparrow$) and top 5 negative ($\downarrow$) correlations shown per persona group. All $p_{\text{FDR}} < 0.005$. Totals: Gender/Feminist (self): 132 significant (27$+$, 105$-$); Other Experts: 119 (26$+$, 93$-$); Lay: 63 (3$+$, 60$-$).}
\label{tab:corr_merged_all}
\end{table*}

\subsection{Per-Model Consistency}
\label{app:gender_per_model}

The direction of the negative associations is consistent across models, though effect sizes vary. For the top negative-correlation terms among Other Expert personas (\textit{all} field), both Qwen3-14B and Gemma-27B show significant accuracy drops when terms are present: \textit{female} (52.5\% vs.\ 69.9\% for Qwen; 54.8\% vs.\ 72.3\% for Gemma), \textit{gender roles} (44.0\% vs.\ 69.7\%; 38.7\% vs.\ 71.6\%), and \textit{gender} (59.6\% vs.\ 69.6\%; 53.1\% vs.\ 71.6\%). GPT-OSS-20B did not reach significance for any of these terms after FDR correction, consistent with its overall low number of significant correlations.





\section{Prompts Collection}
\label{sec:appendix_prompts}

\subsection{Structured Generation}

For LLM persona predictions, we concatenate system prompts with base instructions and structured output enforced by the vllm \cite{kwon2023efficient} inference engine containing the following keys:

\begin{itemize}[noitemsep]
    \item Themes: 2-3 themes interpreted from the text
    \item Key points: The most important narrative points of the text
    \item Evidence: Textual snippets supporting the analysis
    \item Confidence: Scores (1-10) for each analysis
    \item Similarity: Similarity scores (1-10) for texts A and B relative to the anchor
    \item Explanation: A rationale explaining the similarity judgment
\end{itemize}

Pydantic specification for generating structured outputs:

\begin{quote}\small
\begin{verbatim}
from typing import List, Literal
from pydantic import BaseModel, Field

class Analysis(BaseModel):
    themes: List[str] = Field(..., 
        description="2-3 
                     themes or 
                     interpretations")
    key_points: List[str] = Field(..., 
        description="Key 
                     narrative points")
    confidence: int = Field(..., ge=1, 
        le=10, 
        description="Confidence 1-10")
    evidence: List[str] = Field(
        default_factory=list, 
        description="Evidence snippets")

class PersonaResponse(BaseModel):
    analysis_anchor: Analysis
    analysis_a: Analysis
    analysis_b: Analysis
    score_a: int = Field(..., ge=1, 
        le=10)
    score_b: int = Field(..., ge=1, 
        le=10)
    explanation: str = Field(..., 
        description="Brief explanation 
                    of similarity scores")
    rationale: str = Field("", 
        description="Why A or B is more 
        similar to anchor")

class JudgeResponse(BaseModel):
    final_decision: Literal["A", 
        "B"] = Field(description="Which 
            story is more similar to 
            anchor")
    final_confidence: int = Field(ge=1, 
        le=10, description="Decision 
        confidence 1-10")
    explanation: str = Field(description=
        "Brief explanation of decision")

\end{verbatim}
\end{quote}

\subsection{System Prompts}

Example system prompt with instructions for structured output generation:
\begin{quote}\small
\begin{verbatim}
You are a Literary Critic analyzing 
narrative similarity.

Rules:
1. Score similarity 1-10 (higher = more 
    similar to anchor)
2. Confidence 1-10 (higher = more confident)
3. Provide 1-3 evidence snippets per story
4. Brief chain_of_thought (CoT) for each 
    analysis
5. Output ONLY the valid JSON object, no 
    other text

\end{verbatim}
\end{quote}

\subsection{User Prompt}

\begin{quote}\small
\begin{verbatim}
user_prompt = f"""Compare these stories:

ANCHOR STORY:
{anchor}

STORY A:
{text_a}

STORY B:
{text_b}

Analyze which story (A or B) is more similar 
    to the Anchor story."""

\end{verbatim}
\end{quote}

\end{document}